\documentclass[11pt]{article}

\usepackage{acl}
\usepackage{titlesec}
\usepackage{graphicx}
\usepackage{amsmath}
\usepackage{amssymb}
\usepackage{fvextra}
\usepackage{fancyhdr}
\usepackage{setspace}
\usepackage{booktabs}
\usepackage{multirow}

\usepackage{microtype}

\usepackage[english]{babel}

\usepackage[utf8]{inputenc}
\usepackage{times}

\title{CoGR-MoE: Concept-Guided Expert Routing with Consistent Selection and Flexible Reasoning for Visual Question Answering}

\author{
  Xiyin Zeng\thanks{~~Equal contribution.} \qquad 
  Yi Lu\footnotemark[1] \qquad 
  Hao Wang\thanks{~~Corresponding author.} \\
  Hong Kong University of Science and Technology (Guangzhou) \\
  \texttt{\{xzeng310, ylu597\}@connect.hkust-gz.edu.cn} \\
  \texttt{haowang@hkust-gz.edu.cn}
}

\begin{document}

\maketitle
\begin{abstract}
Visual Question Answering (VQA) requires models to identify the correct answer options based on both visual and textual evidence. Recent Mixture-of-Experts (MoE) methods improve option reasoning by grouping similar concepts or routing based on examples.  However, unstable routing can lead to inconsistent expert selection in the same question type, while overly stable routing may reduce flexibility.  To address this, we propose Concept-Guided Routing framework (CoGR-MoE), which incorporates semantics of the answer options to guide expert selection in the training phase.
Next, option features are used to reweight the selected experts, producing discriminative representations for each candidate option.   These option-level representations are further used for option comparison and optimized via contrastive learning. The experimental results indicate that CoGR-MoE delivers strong performance across multiple VQA tasks, demonstrating the effectiveness of our approach. Our code is available at \href{https://github.com/xshjs/CoGR-MoE}{CoGR-MoE}.

\end{abstract}

\section{Introduction}

VQA is a multimodal reasoning task where a model should identify the correct answer among several candidates by grounding its decision in visual and textual evidence\cite{10.1145/3728635}. Flexible multimodal reasoning is essential for real-world applications including visual assistants, robotics and accessibility tools\cite{borisova-etal-2025-scivqa}. However, such fine-grained reasoning remains challenging due to the diversity and complexity of visual–language representations. Recent advances incorporate MoE architectures into vision–language models to allow different experts to specialize in distinct visual or textual patterns\cite{le2025mixtureexpertsmeetspromptbased, nakamura2025dropupcycling}.

The effectiveness of MoE models heavily depends on routing decisions, which can become inconsistent for inputs that require the same answer but are expressed differently\cite{olson2025probingsemanticroutinglarge}.    Recent work has begun to address this routing instability by improving routing mechanisms. Some methods like MoKE\cite{cheng-etal-2025-serial} assign similar concepts to the same knowledge expert, while other approaches\cite{li2025experttokenresonancemoebidirectional} like ERMoE\cite{cheng2025ermoeeigenreparameterizedmixtureofexpertsstable} routes inputs based on their similarity to the experts’ representations. Although these methods maintain routing stability across questions of the same type, such stability also solidifies the activated expert set, thereby weakening flexibility in option-level discrimination\cite{duanmu2025mxmoemixedprecisionquantizationmoe,li2025r2t2reroutingtesttimemultimodal}. As a result, the model struggles to capture the distinct evidence required by each option, leading to errors in comparative reasoning.

Expert roles in MoE models emerge through repeated activation under consistent semantic conditions, requiring cues that remain stable across different representations.    Once the Top-$K$ expert set is static, any remaining discriminative capacity come from how expert outputs are utilized. Accordingly, dynamically adjusting the relative contributions of experts can help improve the model’s ability to distinguish among candidate answers.

\begin{figure*}[t]
    \centering
    \includegraphics[width=\linewidth]{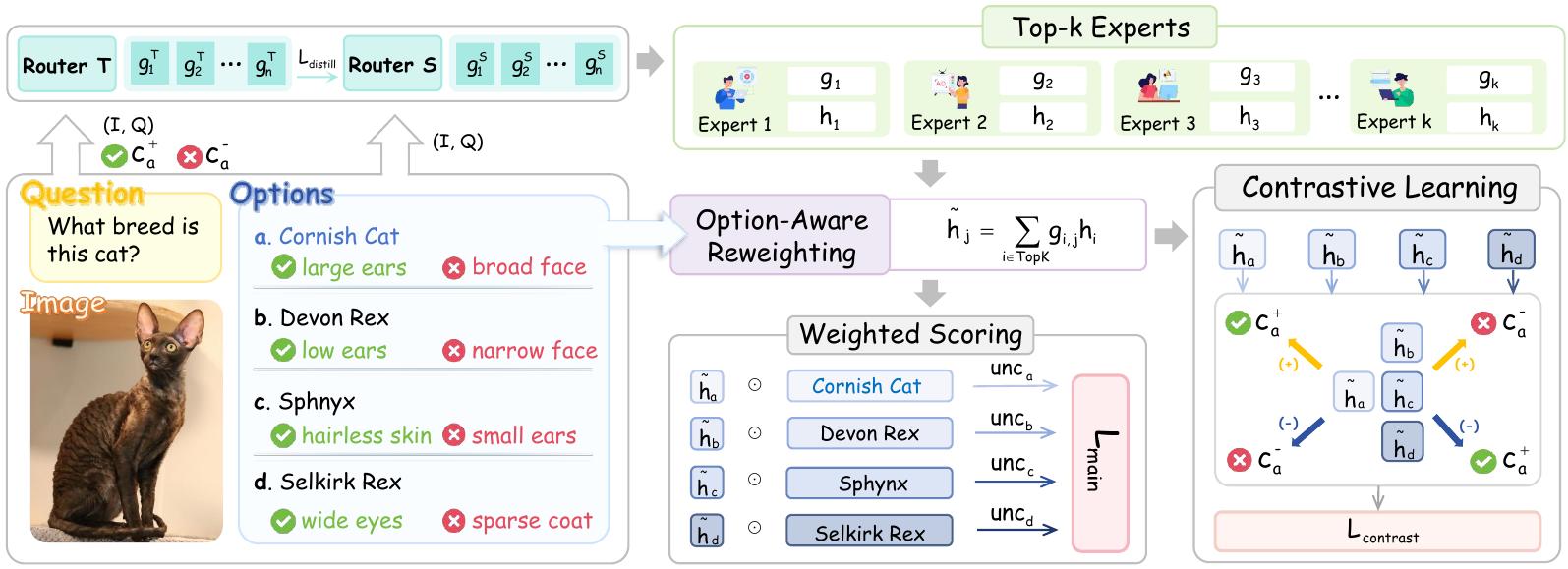}
    \caption{The CoGR-MoE framework incorporates features of the correct answer $c_a^{+}$ and $c_a^{-}$ into the router.
    This process also produces teacher gating weights $g^{T}$ from Router $T$, which are distilled into student gating weights $g^{S}$ of Router $S$. It then applies semantics of option $j$ to flexibly modulate expert contributions $g_{i, j}$, where $i$ indexes the selected experts. The expert outputs $h_i$ are aggregated using gating weights $g_{i, j}$, yielding features $\tilde{h}_j$ for each option. Contrastive learning is employed with positive and negative option pairs $c_a^{+}$ and $c_a^{-}$ to identify correct options.}
    \label{fig:framework}
\end{figure*}

Guided by the above considerations, we propose CoGR-MoE, as illustrated in Figure~\ref{fig:framework}, to mitigate the imbalance between routing instability and expert utilization rigidity. First, positive and negative feature descriptions are generated by a Large Language Model (LLM) for each option. The features of the correct answer are injected into the MoE gating logits as a conceptual direction. After the Top-$K$ experts are selected, feature descriptions from options are used to reweight expert contributions, enabling distinct representations for each option. These option representations are optimized with contrastive learning in the training phase. The experiments show that CoGR-MoE outperforms several strong baselines, indicating the benefit of balancing routing stability with flexible expert utilization.

Our contributions are summarized as follows:
\begin{itemize}
    \item  We propose CoGR-MoE, a concept-guided MoE framework that stabilizes expert selection by injecting answer-relevant features into the routing process.

    \item CoGR-MoE also adjusts expert contributions by option content, enhancing option-level discrimination while preserving routing consistency.
    
    \item Extensive experiments on datasets validate the efficacy of CoGR-MoE, achieving up to a 4.9\% absolute gain in accuracy, with more interpretable expert utilization.
\end{itemize}

\section{Related Works}

\subsection{Embedding-Driven Routing}
Embedding-driven routing computes expert scores from an input’s hidden representation, and typically relies on representation-driven mechanisms to stabilize outputs\cite{li2025unimoe20omniscalinglanguagecentricomnimodal}.
MoME \cite{shen2024momemixturemultimodalexperts} partitions experts into modality-specific and shared groups, making inputs from the same modality more likely to activate the same experts.
In parallel, MH-MoE  \cite{huang2024mhmoemultiheadmixtureofexperts}routes multiple sub-representations of a token to different experts and aggregates their outputs, yielding stable expert combinations despite routing variability. In MMOE\cite{shen2024mome}, similar inputs mapped to nearby points in the joint embedding space are assigned similar expert weights. However, a limitation is that routing is based solely on joint embeddings, without distinguishing modality interactions or semantic roles, which may reduce representation alignment and interpretability in expert selection.

\subsection{Interaction-Aware Routing}
To address the weakened correspondence between routing decisions and cross-modal structure, interaction-driven gating leverages task-specific and cross-modal interaction features to guide expert selection\cite{Cai_2025}\cite{han2025guidingmixtureofexpertstemporalmultimodal}.     I$^{2}$MoE\cite{xin2025i2moe} select expert on interaction types, such that inputs exhibiting the same modality interaction patterns tend to activate the same group of experts.  Similarly, MOEMOE~\cite{verma2025moemoequestionguideddense} guide expert selection with query-aware cross-modal attention, thereby inputs with comparable modality relevance patterns tend to activate similar experts.  RoE-LLaVA\cite{wu2025routingexpertslearningroute} applyies a contrastive objective that pulls semantically related samples closer in the routing embedding space, encouraging them to share relevant expert routes.
These methods mitigate routing shifts caused by interactions, but do not address routing inconsistency when semantically similar inputs are mapped to different embeddings\cite{mu2025comprehensivesurveymixtureofexpertsalgorithms}.

\subsection{Semantic-Informed Routing}
To further mitigate inconsistent expert assignment under representation variation, semantic-informed routing aligns expert selection across semantically related samples.
MoKE\cite{cheng-etal-2025-serial} uses projection-based routing to consistently assign knowledge edits involving similar concepts to the same group of domain-specific experts, while R2-T2~\cite{li2025r2t2reroutingtesttimemultimodal} adjusts routing by comparing new samples with similar past samples.
ProMoE~\cite{wei2025routingmattersmoescaling} introduces a two-step routing mechanism that first separates tokens by functional roles and then assigns them to experts based on similarity to learnable vectors. These methods are built on the mechanism that similar inputs should be routed to the same or similar sets of experts \cite{li2025routingmanifoldalignmentimproves}. 
As a result, the model tends to reuse the same experts even when the question undergoes subtle variations, which can misjudge options.

Routing-based methods enforce concept alignment structurally by controlling expert activation. In contrast, prior VQA research enhances concept-grounded reasoning at the data or objective level, encouraging models to rely on decision-relevant visual concepts rather than language priors\cite{goyal2017makingvvqamatter}. These works enhance concept-level discrimination through stronger grounding signals or counterfactual training\cite{DBLP:journals/pami/ChenZNZX23, wen-etal-2023-digging}.

\section{Methodology}
\subsection{Framework Overview}
Existing MoE routing methods\cite{cheng-etal-2025-serial, li2025routingmanifoldalignmentimproves} achieve more consistent expert selection within the same problem type through semantic grouping. However, this stability can cause routing to become static, repeatedly selecting the similar experts even when the question varies, leading to incorrect answer selection.
To address this limitation, we propose CoGR-MoE, which incorporates  features of answers option into the gating mechanism to align expert selection with the correct answer.  Meanwhile, CoGR-MoE reweights the selected experts using features of options, improving discrimination and reducing incorrect answer selection.

\begin{figure*}[t]
    \centering
    \includegraphics[width=\linewidth]{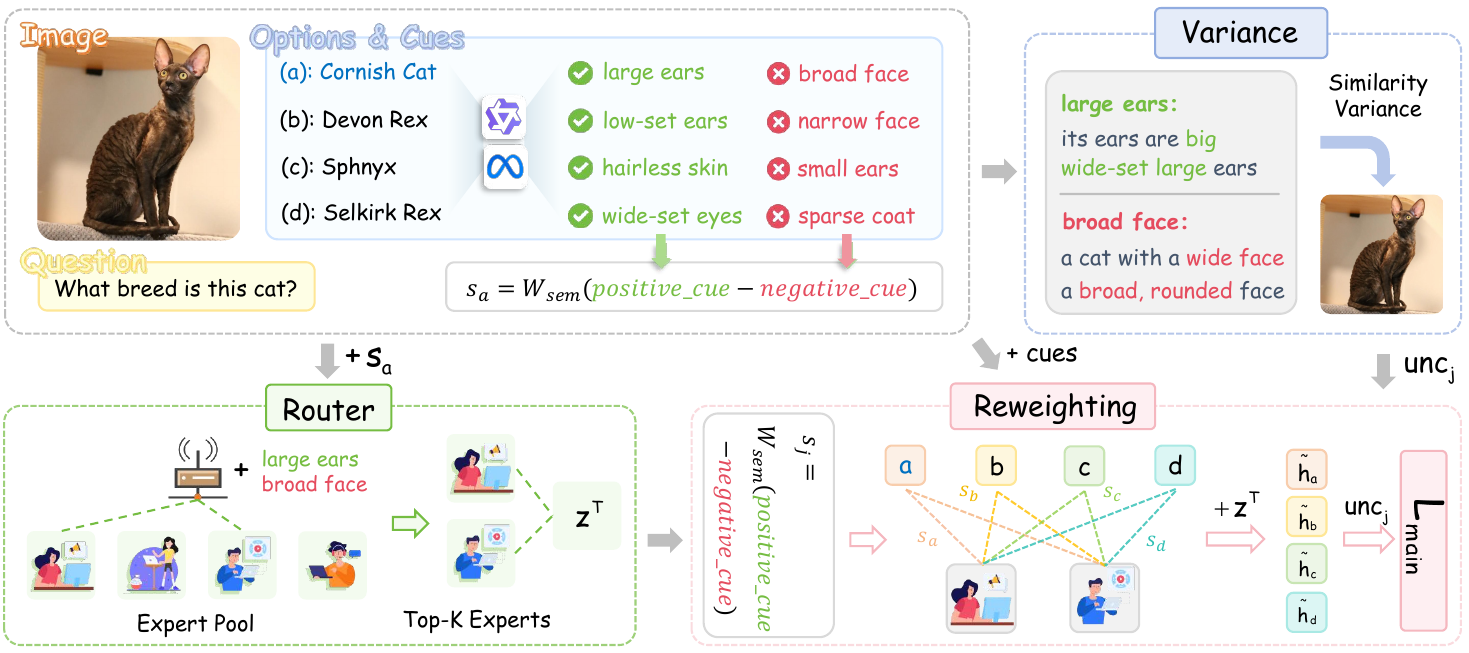}
    \caption{Features of answer option $s_a$ are injected into the router to guide consistent Top-$K$ expert selection, producing initial routing logits $z^{T}$. The uncertainty estimate $\mathrm{unc}_j$ measures the reliability of the $s_j$, which means the features of each option. Together, $s_j$ adjust expert weights to form the option representation $\tilde{h}_j$. $L_{main}$ applies $\mathrm{unc}_j$-weighted cross-entropy to option representations $\tilde{h}_j$.}
    \label{fig:2}
\end{figure*}

In CoGR-MoE, the LLM generates positive and negative visual cues for each option, indicating which attributes must or must not appear for the choice to be correct. The correct answer’s cues are then used to inject into the gating logits to guide the selection of Top-$K$ expert set, as illustrated in Figure~\ref{fig:2}. Then each option uses its own cues to modulate the expert weights, reweighting the selected experts to form an option-specific representation. Finally, representations are used for answer prediction and optimized with a cue-based contrastive loss in the training phase.

\subsection{Semantic Cues Generation}

To provide stable evidence descriptions for routing, an LLM generates positive and negative semantic cues for each answer choice. Positive cues describe what must appear in the image for an option to be correct, while negative cues specify what must not. To quantify how well an option’s cues align with the image, we introduce an agreement score $Agr$. $Agr$ combines the image’s similarity to positive cues with an inversely weighted similarity to negative cues. In parallel, paraphrased variants of the cues are generated via synonym substitutions, and their cue-image similarities are used to compute a variance score ${Var}_j$. Finally, the uncertainty of option \(j\) is defined as a normalized ratio of ${Var}$ to its ${Agr}$:
\begin{equation}
{unc} = \frac{Var}{1 + {Agr}}.
\end{equation}

When ${Var}$ is high, the uncertainty increases because the cues exhibit unstable alignment with the image. Conversely, when ${Agr}$ is high, the uncertainty decreases since the answer is strongly supported by the image semantics. For the answer option, if the estimated uncertainty ${unc}$ exceeds a predefined threshold, semantic cues are regenerated to maintain reliable cue–image alignment.

\subsection{Concept-Guided Expert Routing}
Cues are more stable than specific phrasing because they capture decision-relevant attributes rather than textual variation. As a result, to inject stable guidance into the routing process, we construct a semantic direction $s_a$ from the cues of the correct option $a$. 
It is obtained by taking the difference between the embeddings of positive cues $c_a^{+}$ and negative cues $c_a^{-}$ to support the correct features while suppressing conflicting attributes.

\begin{equation}
s_a = W_{\mathrm{sem}}(c_a^{+} - c_a^{-}).
\label{eq:answer_direction}
\end{equation}

The base logits $z_{\mathrm{base}}$
are the expert scores produced by the MoE router using only the representation of image–question. The semantic direction $s_a$ is added to $z_{\mathrm{base}}$ with a fixed routing strength $\lambda_a = 0.5$ to obtain the concept-guided gating distribution:

\begin{equation}
g^{T} = \mathrm{softmax}\!\left( z_{\mathrm{base}} + \lambda_a\, s_{a} \right).
\end{equation}
where $g^{T}$ denotes the gating distribution produced by the teacher router over the Top-$K$ experts. The teacher and student routers are denoted as Router $T$ and Router $S$, respectively.
The Top-$K$ experts selected by $g^{T}$ are encouraged to align with the semantic direction implied by the cues of the correct option. To enable cue-free inference, Router $S$ with the same architecture operates without $s_a$. Its gating distribution is computed as

\begin{equation}
g^{S} = \mathrm{softmax}( z_{\mathrm{base}} ),
\end{equation}
where $g^{S}$ denotes the gating distribution of Router $S$. This process does not require any cue generation. During training, a KL-divergence loss is introduced to distill the routing distribution of Router $T$ into the Router $S$:

\begin{equation}
L_{{distill}}
=
\mathrm{KL}\!\left(
g^{T} \,\|\, g^{S}
\right).
\label{eq:distill}
\end{equation}

This distillation encourages the Router $S$ to approximate the behavior of Router $T$, enabling it to select experts along the answer-relevant direction at inference without requiring any cues.

\subsection{Option-Aware Expert Reweighting}

After the shared Top-$K$ experts are selected, an option-specific allocation step is required to maintain discriminability among answer choices.
To achieve this, features of each option $j$ are derived from its positive and negative cues to compute $s_j$, which is constructed in the same manner as the routing guidance in Eq.~\eqref{eq:answer_direction}. 
$s_{i, j}$ is injected into the logits of the Top-$K$ experts $i$, adjusting their weights so that experts more relevant to the current option receive higher importance. $z^{T}_i$ denotes the $i$-th component of the Router $T$ routing logits $z^{T}$ before option-aware modulation.

\begin{equation}
{g}_{i, j}
=
\operatorname{softmax}_{i \in \mathrm{TopK}}
\!\left(
z^{T}_i + \lambda_o\, s_{i, j}
\right),
\end{equation}

where $g_{i, j}$ is defined as the $i$-th element of $g_j$, representing the option-specific gating distribution over the Top-$K$ experts $i$. $\lambda_o$ controls the strength of reweighting and is fixed to $0.5$.
Each option then forms its own aggregated representation $\tilde{h}_j$ by weighting the shared expert outputs according to its gating distribution $g_{i, j}$:
\begin{equation}
\tilde{h}_j = \sum_{i \in \mathrm{TopK}} g_{i, j}\, h_i,
\end{equation}
where $i$ denotes the expert index within the Top-$K$ routed experts, and $h_i$ is the output of the $i$-th expert. This mechanism allows all options to share one expert set while still producing distinct representations. If each option were allowed to reweight all experts, the routing would drift and become unstable. Finally, the predictive score $score_j$ for option $j$ is computed as the cosine similarity between its aggregated representation $\tilde{h}_j$ and the embedding of its option text.

\subsection{Training and Inference}
The $L_{\text{main}}$ relies on $score_j$ for each option. However, because different options exhibit different levels of uncertainty, each option is weighted according to the confidence of its cues. $ y_j$ is the one-hot label indicating whether option $j$ is the correct answer:
\begin{equation}
L_{main} = \sum_j \frac{1}{1+{unc}_j}\,
\mathrm{CE}({score_j}, y_j),
\end{equation}
where CE denotes the cross-entropy loss. To further keep the model’s internal representations aligned with decision-relevant semantics, we incorporate a cue-guided contrastive loss. The contrastive loss aligns the aggregated representation of the correct option with its positive cues, while encouraging the representations of incorrect options to be closer to the negative cues:
\begin{equation}
\begin{aligned}
L_{contrast}
&=
-\, \lambda_c \Big[
\cos(\tilde{h}_{\text{correct}},\, c_a^{+})
\\
&\quad
-
\cos(\tilde{h}_{\text{wrong}},\, c_a^{-})
\Big],
\end{aligned}
\end{equation}

where $\tilde{h}_{\text{correct}}$ denotes the representation of the correct option, and $\tilde{h}_{\text{wrong}}$ is computed as a ${score}_j$-weighted average over incorrect options. $\lambda_c$ is a constant value of 0.3 in all experiments.
Finally, the complete training objective integrates all three components, including the $L_{{distill}}$ in  Eq.~\eqref{eq:distill} into a unified optimization target:
\begin{equation}
L_{total}
=
L_{main}
+
L_{contrast}
+
L_{distill}.
\end{equation}

During the training phase, we construct a balanced multiple-choice supervision set by sampling 5,000 questions from each of IconQA\cite{lu2022iconqanewbenchmarkabstract}, A-OKVQA\cite{schwenk2022aokvqabenchmarkvisualquestion}, and Visual7W\cite{zhu2016visual7wgroundedquestionanswering}.
The same training questions are used for all baselines to ensure a fair comparison. 

During the inference phase, no cues are generated. CoGR-MoE relies solely on the Router $S$, which selects the Top-$K$ experts based on the image–question pair. These experts are computed once and shared across all candidate answers. For each option $j$, the learned option-specific adjustment signal obtained from its text embedding is applied to allocate weights over the shared experts, producing an aggregated representation $\tilde{h}_j$. The model then computes a cosine similarity score between $\tilde{h}_j$ and the option text embedding, and the option with the highest score is chosen as the final answer.

\begin{table*}[t]
\centering

\caption{Performance comparison of CoGR and other different MoE-based models on the MRAG-Bench.  Results are reported for overall accuracy and across sub-categories including Perspective, Transformative, and Others. Detailed results for finer-grained sub-categories are provided in the Appendix~\ref{sec:appendix}.}
\begin{tabular}{l|c|c|c|c}
\toprule
\textbf{Methods} & \textbf{Overall} & {\textbf{Perspective}} & {\textbf{Transformative}} & \textbf{Others} \\
\midrule
\multicolumn{5}{l}{\textit{MOE-LLaVA}\cite{lin2024moe} }\\
\midrule
MOE-LLaVA & 53.29 & 55.54 & 50.78  & 51.67 \\
MH-MoE \cite{huang2024mhmoemultiheadmixtureofexperts}& 61.71 & 69.78 & 51.70 & 40.83 \\
Metis-HOME \cite{lan2025metis} & 57.25 & 63.79 &51.10 & 55.75 \\
I$^{2}$MoE \cite{xin2025i2moe} & 59.35 & 62.16 & 50.87 & 60.83 \\
CL-MOE \cite{huai2025cl} & 58.95 & 63.15 & 51.57 & 59.94 \\
MoME \cite{shen2024momemixturemultimodalexperts}& 62.76 & \textbf{69.30} & 55.91 & \textbf{64.06}\\
CoGR-MoE (Ours)& \textbf{63.25} & 69.09 & \textbf{58.55} & 58.67 \\

\midrule
\multicolumn{5}{l}{\textit{Qwen3-VL-A3B-30B} \cite{yang2025qwen3technicalreport}}\\
\midrule

Qwen3-VL-A3B-30B & 59.32 & 64.08 & 54.29 & 61.33 \\
MH-MoE & 64.85 &  65.92 & 58.82 & 50.50 \\
Metis-HOME &65.81 & 68.45 & 61.40 &65.30\\
I$^{2}$MoE & 64.27 & 68.71 &53.46 & \textbf{68.00} \\
CL-MOE & 65.31 & 73.30 & 48.69 & 48.65 \\
MoME & 66.78 &   70.64 & 56.39 & 56.75\\
CoGR-MoE (Ours)& \textbf{68.96} & \textbf{74.54}  & \textbf{64.61} & 62.83 \\

\bottomrule
\end{tabular}
\label{tab:mrag_simplified}
\end{table*}

\begin{table*}[ht]
\centering
\caption{Comparison of MoE-based LLaVA-1.5-7B \cite{liu2024improvedbaselinesvisualinstruction} on five standard VL benchmarks in VMCBench, reporting accuracy across VQAv2\cite{goyal2017makingvvqamatter}, GQA\cite{hudson2019gqanewdatasetrealworld}, VizWiz\cite{gurari2018vizwizgrandchallengeanswering}, ScienceQA\cite{lu2022learnexplainmultimodalreasoning}, MMVet\cite{yu2024mmvetevaluatinglargemultimodal}, and MMStar\cite{chen2024rightwayevaluatinglarge}.}
\begin{tabular}{l|cccccc}
\toprule
\textbf{Method} & \textbf{VQAv2} & \textbf{GQA} & \textbf{VizWiz} & \textbf{ScienceQA}  & \textbf{MMVet} & \textbf{MMStar} \\
\midrule
LLaVA-1.5-7B & 66.7  & 72.6 &68.6 &64.3 & 54.0 & 34.2 \\
MH-MoE &  83.6 & 75.3 & 77.0 & 77.4 & 64.0 & 39.9 \\
I$^{2}$MoE &  83.3 & 81.1 & 82.4 & 74.6 & \textbf{66.7} & 47.8 \\
CL-MOE     & 79.2 & 78.2 & 80.9 & \textbf{83.0} & 60.4 & 42.8\\ MoME       & 75.7 & 81.2 & 81.0 & 80.4 & 63.3 & 48.1  \\
Metis-HOME & 82.7 & \textbf{83.2} & 78.5 & 80.5 & 64.7 & 50.4\\
CoGR-MoE (Ours) & \textbf{88.5} & \textbf{83.2} & \textbf{84.8} & 77.4 &  64.3 & \textbf{52.0} \\
\bottomrule
\end{tabular}
\label{tab:a1}
\end{table*}

\section{Experiments}
\subsection{Experimental Setup}

We evaluate CoGR-MOE on VMCBench\cite{AutoConverter} and MRAG-Bench\cite{hu2024mragbench}, two benchmarks for vision-language reasoning, and compare it with other MoE methods. VMCBench unifies twenty existing VQA datasets by converting open-ended questions into four-option multiple-choice, while MRAG-Bench is a vision-centric benchmark with images and annotated multiple-choice questions across several scenarios. Results are evaluated using task accuracy as the performance metric.

To further quantify the contribution of each component, we conduct an ablation study in which each variant disables one module while keeping remaining modules unchanged. Variants of CoGR-MoE include:

\begin{itemize}
    \item \textbf{w/o $s_{\text{a}}$:} removes the correct-answer semantic $s_{\text{a}}$ and forms the routing anchor solely from image–question pair, eliminating explicit semantic grounding.
    \item \textbf{w/o $s_j$:} keeps the Top-$K$ experts but removes the option-specific semantic term $s_j$, forcing all options to share the same routing-level expert weights, with option discrimination relying only on option text embeddings.
    \item \textbf{w/o ${unc}_j$:} removes uncertainty-aware weighting derived from cue consistency, and applies uniform weighting across all samples during training.
    \item \textbf{w/o  $L_{{contrast}}$:} removes the contrastive loss that aligns expert representations with positive answer semantics while pushing them away from negative cues.
    \item \textbf{w/o $\mathcal{L}_{{distill}}$:} removes the distillation loss that transfers soft routing signals from the Router $T$ to the Router $S$.
    \item \textbf{Prompt-only:} removes all cue-based training components, and uses the same LLM prompt only at inference phase to isolate the effect of cue-based training.
\end{itemize}

The influence of the MoE architecture on CoGR-MoE is examined by varying the total number of experts $n \in \{1,2,4,8\}$ and the number of activated experts $K \in \{1,2,3,4\}$ on LLaVA-1.5-7B. Each $(n,K)$ configuration is evaluated using two metrics. Accuracy reflects practical performance, while the semantic–expert alignment score $\mathrm{Sim}$ measures how well expert routing follows the intended semantic direction.
\begin{equation}
\mathrm{Sim}
=
\cos\!\left(
h_{\mathrm{Top}K},\,
s_a
\right),
\end{equation}
where $s_a$ denotes the semantic direction of the correct answer, and 
$h_{\mathrm{Top}K}$ represents the aggregated representation of the Top-$K$ experts.

\begin{figure*}[t]
    \centering
    \includegraphics[width=\linewidth]{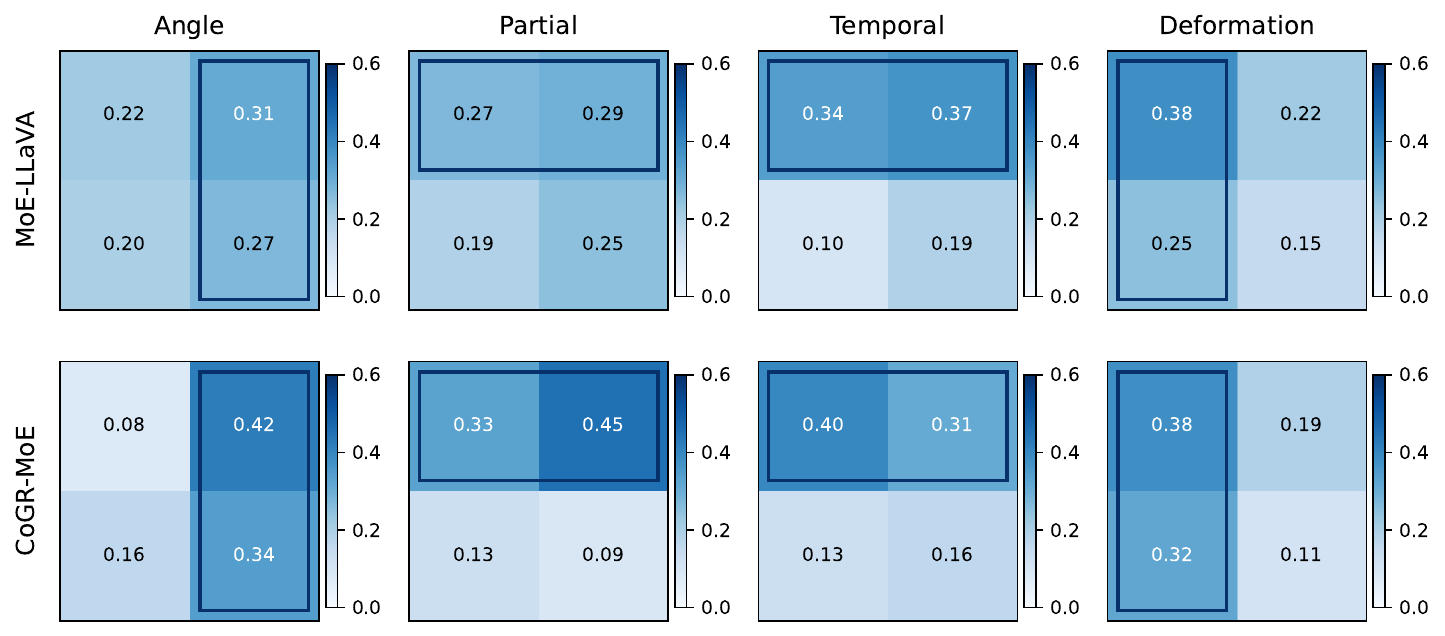}
    \caption{Each heatmap summarizes expert routing behavior of MoE-LLaVA and CoGR-MoE under four subtasks in MRAG-Bench. Each cell corresponds to a single expert. Color intensity indicates the probability of an expert being selected into the Top-$K$ routing set. Darker colors correspond to experts that are more frequently selected under the given task type. Compared to MoE-LLaVA, CoGR-MoE exhibits more concentrated and task-consistent routing patterns.  }
    \label{fig:3}
\end{figure*}

\subsection{Analysis}

As shown in Table~\ref{tab:mrag_simplified}, CoGR-MoE yields the highest overall accuracy on two backbones, reaching 63.25 on MOE-LLaVA and 68.96 on Qwen3-VL-A3B-30B. It shows notable gains in Perspective and Transformative tasks, highlighting its strength in handling partial-visibility and spatial transformations. However, improvements on the Others subset remain limited compared with I$^{2}$MoE and MoME. CoGR-MoE demonstrates superior performance across the VMCBench in Table~\ref{tab:a1}, particularly excelling in tasks like VQAv2 and VizWiz. However, its performance on the ScienceQA and MMVet tasks shows less improvement, lagging behind CL-MoE and Metis-HOME.

The ablation results in Table~\ref{tab:abla1} highlight the contribution of each component to the overall performance of CoGR-MoE.  On almost all datasets, the Full model still performs better.  Notably, the w/o ${unc}_j$ variant outperforms the Full model on ScienceQA which involves more direct reasoning, achieving a score of 78.0, compared to the Full model's 77.6. In addition, the removal of $s_{\text{a}}$ and $ L_{\text{distill}}$, as well as the Prompt-only variant, had the largest negative impact on accuracy, causing a significant drop in performance across multiple datasets.

In Table~\ref{tab:topk}, the best results occur at $n=8$ when $n$ changes, reaching an accuracy of 71.6 and a $\mathrm{Sim}$ of 0.48, noticeably higher than smaller settings. 
When varying $K$ under $n=8$, activating $K=2$ yields the highest accuracy of 74.5 and the strongest $\mathrm{Sim}$ of 0.51, whereas larger $K$ values lead to a decline.

\begin{figure}[t]
    \centering
    \includegraphics[width=1\columnwidth]{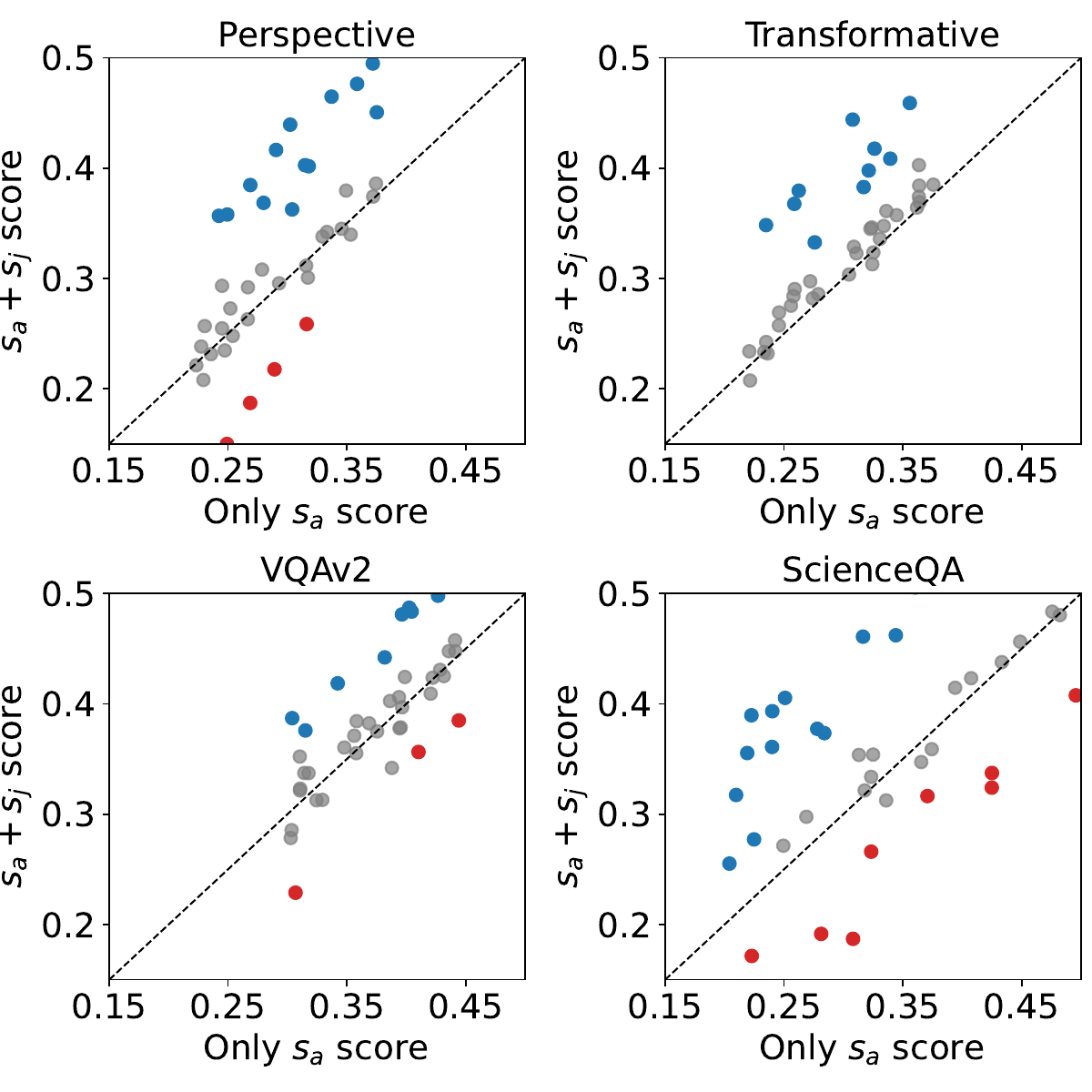}
    \caption{Comparison of answer scoring with and without option-level reweighting. The top row shows subtasks Perspective and Transformative in MRAG-Bench and the bottom row shows VQAv2 and ScienceQA in VMCBench, each with 40 randomly sampled instances.  The x-axis shows scores computed using only $s_a$, whereas the y-axis shows scores of $s_a + s_j$. Blue and red points indicate notable score increases and decreases, respectively, while gray points indicate minor changes.}
    \label{fig:semantic-alignment}
\end{figure}

\begin{table*}[ht]
\centering
\caption{Ablation results for the components built upon the MoE-LLaVA backbone. }
\begin{tabular}{l|ccccccc}
\toprule
\textbf{Method} & \textbf{VQAv2} & \textbf{GQA} & \textbf{VizWiz} & \textbf{ScienceQA}  & \textbf{MMVet} & \textbf{MMStar} & \textbf{MRAG} \\
\midrule
w/o $s_{\text{a}}$ & 80.4  & 76.8 &80.4 & 70.1& 63.2 &46.3 & 57.9 \\
w/o $s_j$ & 84.6 & 82.3 & 82.7 & 76.1 & 64.1 & 50.9 & 60.8 \\
w/o ${unc}_j$ &  85.2& 80.8& 83.5& \textbf{78.0}& 62.5& 50.8& 61.0 \\
w/o $ L_{{contrast}}$ &  87.8 & 84.6   & 82.1&73.0 &60.4 &49.5 &59.3 \\
w/o $L_{{distill}}$ &  83.3 & 82.4 & 79.8 & 75.8 &  63.2 & 50.4 & 58.9\\
Prompt-only &  81.5 & 77.6 & 79.4 & 65.7 &  63.6 & 45.1& 57.3\\
Full & \textbf{88.2} & \textbf{85.1} & \textbf{85.8} & 77.6 &  \textbf{65.7} & \textbf{53.4} & \textbf{63.3}\\
\bottomrule
\end{tabular}
\label{tab:abla1}
\end{table*}

\begin{table}[t]
\centering
\caption{
Effect of the number of experts $n$ and activated experts $K$ on accuracy and $\mathrm{Sim}$ on VMCBench.
When varying $n$, the number of activated experts is fixed to $K=1$; when varying $K$, the total number of experts is fixed to $n=8$.}
\begin{tabular}{c | c c c | c c c}
\toprule
Method & $n$ & Acc & $\mathrm{Sim}$ & $K$ & Acc & $\mathrm{Sim}$ \\
\midrule
\multirow{5}{*}{CoGR-MoE}
& 1  & 68.2 & 0.36 & 1 & 71.6 & 0.48 \\
& 2  & 64.5 & 0.27 & 2 & \textbf{74.5} & \textbf{0.51} \\
& 4  & 59.1 & 0.43 & 3 & 68.6 & 0.38 \\
& 8  & \textbf{71.6} & \textbf{0.48} & 4 & 64.2 & 0.44 \\
& 16 & 55.4 & 0.30 & 5 & 62.0 & 0.31 \\
\bottomrule
\end{tabular}
\label{tab:topk}
\end{table}

\subsection{Discussion}
CoGR-MoE excels because it learns to select experts based on correct semantic direction and consistently assigns similar questions to the same experts. Figure~\ref{fig:3} shows that consistent routing of similar questions promotes reuse of experts aligned with content requirements.
To further support fine-grained distinctions among answer options, option-specific cues are introduced to dynamically reweight the shared Top-$K$ experts. This enables option-level discrimination without disrupting established expert roles, as illustrated in Figure~\ref{fig:semantic-alignment}. Even when the initially selected experts provide weakly discriminative scores, reweighting can still amplify relative differences across options.
This effect is most pronounced in Perspective and Transformative tasks involving viewpoint changes or partial visual evidence.  In contrast, gains on VQAv2 are modest, while ScienceQA shows more unstable effects due to higher uncertainty.

However, CoGR-MoE underperforms in the task like Other because it often fails to capture the subtle differences in cross-modal interactions.
The core limitation lies in that its routing strategy is primarily guided by the overall semantic direction of the input, rather than explicitly modeling fine-grained cross-modal interactions.
Concept-guided gating applies a consistent expert-selection strategy aligned with the overall semantic direction of the input, whereas interaction-driven gating adapts expert selection to the instance-specific cross-modal interactions required by each sample.
In tasks that demand highly dynamic and complex multimodal interactions, routing based primarily on semantic alignment may be insufficient to capture the full range of instance-specific dependencies.
In contrast, I²MoE excels by using interaction-specific experts for each modality pair, allowing more precise expert selection in tasks with complex intermodal relationships. Similarly, MoME performs well by dynamically activating expert pools specialized for visual and textual tasks.

Prompt-only and the w/o $s_{\text{a}}$ variants underperform because inference-time prompting or unguided routing does not update the router. By contrast, cue-based training internalizes attribute–expert associations through repeated gradient reinforcement.  In addition, removing the contrastive loss disrupts CoGR-MoE’s ability to generate more discriminative expert representations. This is especially important in tasks like VQAv2 and GQA, which require fine-grained reasoning. In the case of ScienceQA, many challenging examples arise from reasoning complexity or from images that lack clear visual evidence for the correct answer.    Uncertainty-based weighting reduces the gradient updates for cue-weak but informative examples, while allowing easier, cue-dominated examples to dominate the gradient updates during optimization.  Under this training scheme, the Full model places less emphasis on cue-weak examples that may be more instructive, which hinder its ability to acquire fine-grained reasoning skills.

Performance peaks at $n=8$, indicating that a balanced expert pool provides sufficient diversity for specialization.
Smaller configurations lack capacity for experts to develop distinct semantic roles, while overly large pools dilute expert utilization.
Under a fixed $n=8$, the optimal setting occurs at $K=2$, where experts can contribute complementary semantic information while maintaining focused routing.
Larger $K$ values introduce semantic dilution, as aggregating many weakly relevant experts reduces both accuracy and semantic alignment.

\section{Conclusion}
In this paper, we propose CoGR-MoE, a concept-guided MoE framework that mitigates routing inconsistency by injecting the correct answer’s semantic direction into router.  Furthermore, the option-aware weighting mechanism dynamically reweights the Top-$K$ experts, enhancing the model’s ability to discriminate.  Experiments on multiple multimodal benchmarks demonstrate the superiority of CoGR-MoE in improving accuracy across diverse VQA tasks.  Ablation studies further confirm the effectiveness of each component.
Future work will focus on extending CoGR-MoE’s routing mechanism to better account for varying cross-modal patterns and task complexities.

\section{Limitations}
Despite its effectiveness, CoGR-MoE has several limitations.
First, it relies on LLM-generated semantic cues during training, introducing additional computational cost and dependency on external language models, which may limit scalability for large-scale or frequent retraining settings.

Second, although the agreement-variance mechanism alleviates unstable or noisy cues, the model may still be sensitive to systematic biases or hallucinations from the LLM, particularly for visually ambiguous or fine-grained concepts.

Finally, our evaluation is limited to multiple-choice visual question answering benchmarks, and the effectiveness of the proposed mechanisms for open-ended generation or other multimodal reasoning tasks remains to be explored.

\section{Acknowledgement}
This work is supported by the National Natural Science Foundation of China (No. 62406267), Guangdong Provincial Project (No. 2024QN11X072) and Guangzhou Municipal Science and Technology Project (No. 2025A04J4070).

\appendix

\section{Training Setup}
\label{sec:appendix}
We fine-tune Qwen3-VL-30B, MoE-LLaVA, and LLaVA-1.5-7B on the IconQA, A-OKVQA, and Visual7W training sets, where 5,000 samples are randomly selected from each dataset. LLaVA-1.5-7B is converted to an MoE model by replacing dense MLP blocks with 8-expert Top-2 gated MoE layers.  For Qwen3-VL-30B and MoE-LLaVA, we follow their official MoE configurations as specified in the original model releases.

We use the AdamW optimizer with a base learning rate of $1\times10^{-4}$ and a per-GPU batch size of 4. Training is conducted on 4 NVIDIA A100 GPUs with gradient accumulation over 4 steps, resulting in an effective batch size of 64. A cosine learning rate scheduler (CosineAnnealingLR) is applied with a minimum learning rate of $1\times10^{-6}$ and a warmup period of 100 steps. Gradient clipping with a maximum norm of 1.0 is used to stabilize training. In addition, sample-level gradient reweighting is applied at the routing layer based on ${Agr}(j)$ and ${Var}(j)$, with weights clipped to the range $[0.1, 1.0]$.

Compared to a standard Top-$K$ MoE architecture, our approach retains the same gating scheme and does not increase the number of expert forward passes during training or inference, introducing only a lightweight semantic term in the routing logits. At inference time, the model exhibits the same inference-time complexity and memory footprint as the corresponding baseline MoE models.

Semantic cues are generated offline using a large language model, with one-time generation per question prior to training.  The generated cues are cached and reused throughout training and evaluation.  As a result, no LLM calls are made during training iterations or inference, and the LLM is not part of the training or inference loop. All reported results are averaged over three independent runs.

\begin{table*}[t]
\centering
\caption{Performance comparison on the \textbf{Perspective} categories of MRAG-Bench.}
\label{tab:mrag_perspective}
\begin{tabular}{l|cccc}
\toprule
\textbf{Methods} & Angle & Partial & Scope & Occlusion \\
\midrule
\multicolumn{5}{l}{\textit{MOE-LLaVA}} \\
\midrule
MOE-LLaVA          & 62.42 & 54.88 & 60.19 & 47.65 \\
MH-MoE             & 68.94 & \textbf{69.92} & \textbf{73.53} & 66.74 \\
Metis-HOME         & 68.77 & 62.68 & 65.24 & 58.47 \\
I$^{2}$MoE         & 65.49 & 61.76 & 64.87 & 59.53 \\
CL-MOE             & 65.69 & 63.41 & 58.82 & 66.67 \\
MoME               & 68.95 & 69.42 & 73.24 & 65.57 \\
CoGR-MoE (Ours)    & \textbf{70.29} & \textbf{69.92} & 67.35 & \textbf{68.81} \\
\midrule
\multicolumn{5}{l}{\textit{Qwen3-VL-A3B-30B}} \\
\midrule
Qwen3-VL-A3B-30B   & 62.80 & 61.94 & 63.78 & 67.81 \\
MH-MoE             & 71.01 & \textbf{73.42} & 57.90 & 61.33 \\
Metis-HOME         & 64.78 & 68.54 & 69.83 & 70.65 \\
I$^{2}$MoE         & 69.53 & 67.29 & 72.61 & 65.42 \\
CL-MOE             & 72.61 & 72.92 & 73.37 & 74.30 \\
MoME               & 74.18 & 72.54 & 66.86 & 68.96 \\
CoGR-MoE (Ours)    & \textbf{77.78} & 71.26 & \textbf{73.61} & \textbf{75.63} \\
\bottomrule
\end{tabular}
\end{table*}

\begin{table*}[t]
\centering
\caption{Performance comparison on the \textbf{Transformative} and \textbf{Others} categories of MRAG-Bench.}
\label{tab:mrag_transformative}
\begin{tabular}{l|cccc|c}
\toprule
\textbf{Methods} & Temporal & Deformation & Incomplete & Biological & Others \\
\midrule
\multicolumn{6}{l}{\textit{MOE-LLaVA}} \\
\midrule
MOE-LLaVA          & 47.65 & 53.92 & 47.62 & 53.92 & 51.67 \\
MH-MoE             & \textbf{70.47} & 42.16 & 41.22 & 52.94 & 40.83 \\
Metis-HOME         & 58.59 & 50.28 & 40.63 & 54.90 & 55.75 \\
I$^{2}$MoE         & 54.94 & 57.69 & 32.06 & 58.80 & 60.83 \\
CL-MOE             & 63.12 & 47.06 & 39.22 & 56.86 & 59.94 \\
MoME               & 69.30 & 54.71 & 39.82 & \textbf{59.80} & \textbf{64.06} \\
CoGR-MoE (Ours)    & 68.46 & \textbf{58.80} & \textbf{49.18} & 57.75 & 58.67 \\
\midrule
\multicolumn{6}{l}{\textit{Qwen3-VL-A3B-30B}} \\
\midrule
Qwen3-VL-A3B-30B   & 57.36 & 51.85 & \textbf{52.02} & 55.94 & 61.33 \\
MH-MoE             & 71.46 & 58.88 & 34.37 & 65.78 & 50.50 \\
Metis-HOME         & 74.26 & 62.67 & 51.94 & 64.32 & 65.30 \\
I$^{2}$MoE         & 62.73 & 60.41 & 42.22 & 60.84 & \textbf{68.00} \\
CL-MOE             & 71.46 & 47.12 & 41.27 & 56.92 & 48.65 \\
MoME               & 73.05 & 56.92 & 41.37 & 63.26 & 56.75 \\
CoGR-MoE (Ours)    & \textbf{74.84} & \textbf{65.21} & 51.62 & \textbf{66.75} & 62.83 \\
\bottomrule
\end{tabular}
\end{table*}

\section{Additional Experiments}

\begin{figure*}[t]
    \centering
    \includegraphics[width=\linewidth]{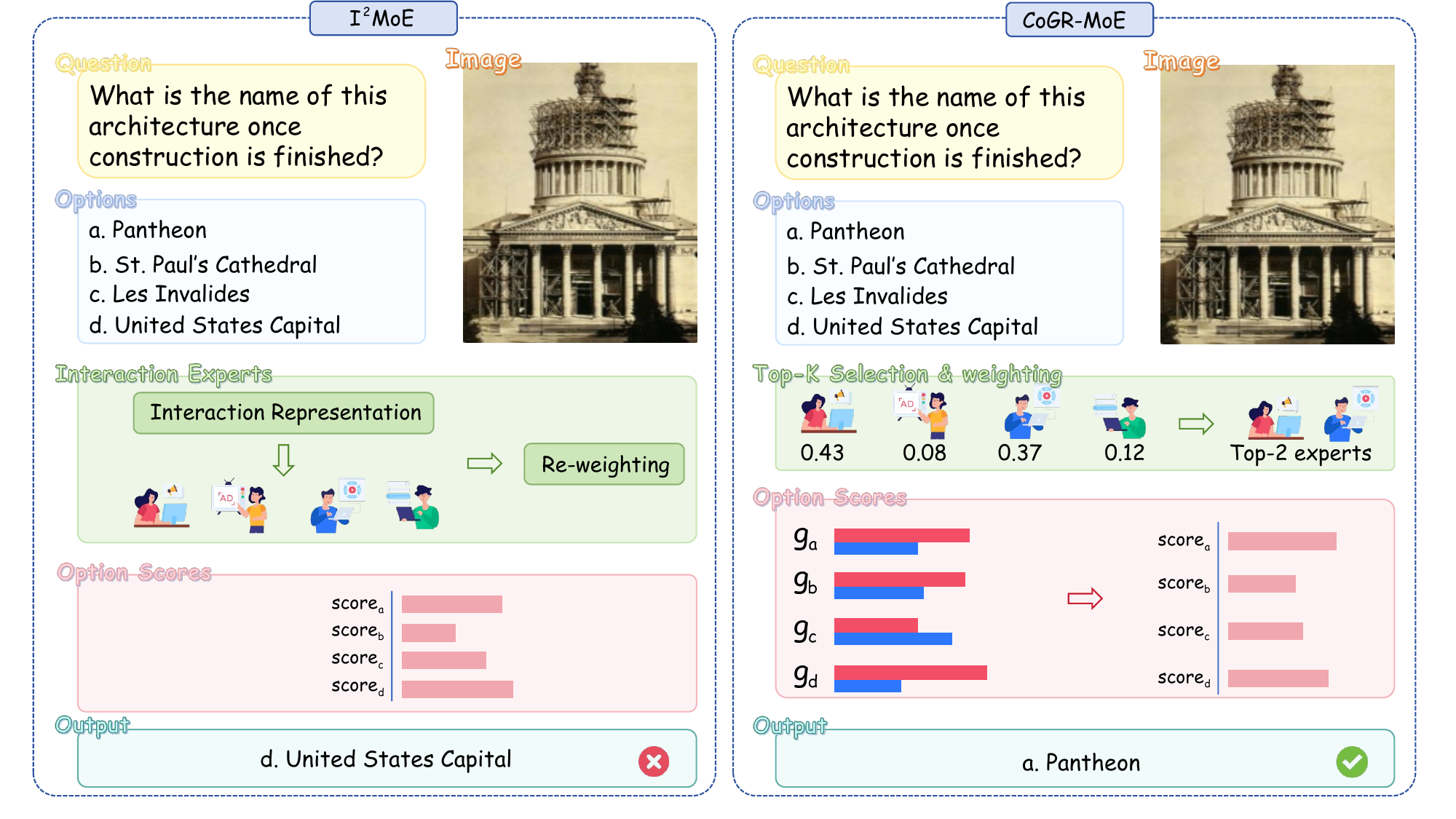}
    \caption{Qualitative comparison between CoGR-MoE and I$^{2}$MoE on a temporal question from MRAG-Bench.  Although both models process the same image–question pair, CoGR-MoE produces more discriminative option scores by leveraging structured expert uCoGR, leading to the correct prediction.}
    \label{fig:6}
\end{figure*}

\subsection{Case Study}

Figure~\ref{fig:6} presents an example from the Temporal category of MRAG-Bench, comparing the behaviors of CoGR-MoE and I$^{2}$MoE. The image depicts a large neoclassical building under construction, while the question asks for the name of the architecture once construction is finished. Among the four candidate answers, the correct choice is Pantheon.

We compare the routing behavior and decision outcomes of MH-MoE and CoGR-MoE on this example. I$^{2}$MoE exhibits a relatively diffuse routing pattern, with expert activations distributed more evenly across the selected experts. As a result, the aggregated representations formed for different answer options are weakly differentiated, leading the model to assign a higher similarity score to an incorrect option. In contrast, CoGR-MoE shows a more structured routing behavior.       Although the same Top-$K$ experts are selected, CoGR-MoE applies option-specific reweighting over these experts, resulting in distinct aggregated representations for each answer choice.

This difference in expert CoGR is reflected in the final option scores. CoGR-MoE substantially increases the score of the correct answer while suppressing the scores of distractors, thereby enlarging the margin between the correct option and all alternatives. I$^{2}$MoE , by contrast, fails to separate the correct answer from visually or semantically similar distractors.

\subsection{Uncertainty Study}

To evaluate the contribution of stability-aware sample weighting, we disable the variance component in uncertainty estimation. As a result, gradient weighting depends solely on agreement, without suppressing unstable or noisy probes. All options therefore receive similar training strength independent of textual consistency. 
To further isolate the role of agreement-based semantic reliability, we construct an Only-Variance baseline by removing the agreement term. In this setting, uncertainty is determined solely by probe variation, without incorporating agreement information.
This variant treats all semantic cues as equally plausible regardless of their semantic alignment with the image–question pair.
It keeps the ability to detect probe inconsistency (larger variance → higher uncertainty) but removes any information about whether the probes actually align with the image–question pair. 

\begin{table*}[ht]
\centering
\small 
\caption{Ablation accuracy on uncertainty components.}
\begin{tabular}{l|ccccccc}
\toprule
\textbf{Method} & \textbf{VQAv2} & \textbf{GQA} & \textbf{VizWiz} & \textbf{ScienceQA}  & \textbf{MMVet} & \textbf{MMStar} & \textbf{MRAG} \\
\midrule

w/o unc &  82.4 & 81.1 & 83.0 & 75.9 & 63.5 & 49.7 & 59.5\\
Only-Variance & 85.8 & 82.5 & 84.2 & 75.4 & 62.8 & 50.9 & 62.7 \\
Full & \textbf{88.5} & \textbf{83.2} & \textbf{84.8} &\textbf{77.4} &  64.3 & \textbf{52.0} & \textbf{63.3}\\
\bottomrule
\end{tabular}
\label{tab:unc}
\end{table*}

Across all seven benchmarks, removing uncertainty consistently lowers accuracy as shown in Table~\ref{tab:unc}.   The largest drops appear on VQAv2, with a decrease of 4.1 percent.   More moderate declines occur on ScienceQA with 1.5 percent and MMVet with 0.8 percent. The accuracy of Only-Variance falls between that of w/o uncertainty and Full CoGR.   However, on ScienceQA, Only-Variance obtains 75.4 percent, slightly lower than the 75.9 percent of w/o uncertainty, and on MMVet, Only-Variance reaches 62.8 percent, also the lowest. 

The largest performance drops on VQAv2 and MRAG stem from the fact that they contain highly diverse, semantically overlapping answer options and substantial linguistic ambiguity, which makes them particularly sensitive to noisy or unstable probe signals.          Without uncertainty, noisy probes receive equal training strength, contaminating expert routing, distorting the semantic direction for contrastive alignment, and amplifying option-level confusion.     ScienceQA, however, is largely unaffected by this removal because its questions are highly structured and semantically unambiguous.   As a result, the model relies far less on statement stability for distinguishing candidate answers, leading to only minimal degradation when uncertainty is removed.

On ScienceQA, the Only-Variance variant underperforms because questions are highly structured, so variance mainly reflects random fluctuations in statement scoring rather than meaningful semantic stability. On MMVet’s instruction- and reasoning-style questions, the performance drop is even more pronounced. Correct answers are often longer, multi-aspect explanations, which naturally induce higher variance across statements, whereas plausible but incorrect distractors can appear more uniform and thus exhibit lower variance. A variance-only scheme therefore tends to down-weight precisely those samples that require nuanced reasoning, while relatively up-weighting simpler but incorrect options.             In contrast, the Full model formulation combines variance with agreement-based alignment, allowing the model to discount probes that are stable yet semantically misaligned.

\subsection{Prompt Ablation Study}
To isolate prompt effects from the CoGR-MoE architecture itself, we compare two cues-generation settings under an otherwise identical training and evaluation pipeline. We consider the following two prompts for the LLM-based cues generator:

\begin{itemize}
    \item  \textbf{Full Prompt.}
This is the task-aware prompt used in our main experiments.      It explicitly instructs the LLM to (i) distinguish between positive-cues and negative-cues, and (ii) attend to task characteristics (e.g., object recognition, counting, OCR-like text reading, and relational reasoning). 

    \item \textbf{Minimal Prompt.}
    This ablated version removes all task-type hints and high-level reasoning instructions.    The LLM is only asked to generate a small set of positive-cues and negative-cues for each answer option.    No additional guidance about problem type, reasoning strategy, or probe diversity is provided. 

\end{itemize}

\begin{table*}[ht]
\centering
\small 
\caption{Performance comparison of CoGR-MoE under different prompt configurations.
The Full Prompt provides detailed visual cue guidance, while the Minimal Prompt only generates simple postive and negative cues.   All other components of CoGR-MoE remain identical.}
\begin{tabular}{l|ccccccc}
\toprule
\textbf{Method} & \textbf{VQAv2} & \textbf{GQA} & \textbf{VizWiz} & \textbf{ScienceQA}  & \textbf{MMVet} & \textbf{MMStar} & \textbf{MRAG} \\
\midrule
Minimal Prompt &  87.8 & 82.6   & 82.1 & 74.3 &60.9 &47.5 &59.6 \\
Full Prompt & \textbf{88.5} & \textbf{83.2} & \textbf{84.8} &\textbf{77.4} &  \textbf{64.3} & \textbf{52.0} & \textbf{63.3}\\
\bottomrule
\end{tabular}
\label{tab:vqa_results}
\end{table*}

Across the seven benchmarks, using the Minimal Prompt instead of the Full Prompt leads to only small drops on the standard VQA datasets.     The decrease is very small on VQAv2, where accuracy falls from 88.5 to 87.8, and similarly small on GQA, where it goes from 83.2 to 82.6.     Larger differences appear on more complex reasoning datasets: MMVet from 64.3 to 60.9, and MMStar from 52.0 to 47.5.    The decline is most pronounced on reasoning-oriented and instruction-following tasks.

The prompt ablation highlights a clear task-dependent pattern.   On perception-oriented VQA datasets such as VQAv2 and GQA, replacing the full task-aware prompt with a minimal cue prompt leads to only minor degradation, suggesting that these benchmarks rely mainly on local visual evidence that simple support/reject cues can already capture.

In contrast, reasoning-oriented and instruction-following tasks, including ScienceQA, MMVet, MMStar, and MRAG show larger drops.   These datasets require understanding multi-step instructions and applying task-specific constraints, and the full prompt provides essential guidance for generating cues that reflect such reasoning steps.   Without this guidance, the cues become more superficial, reducing the quality of semantic anchoring and routing. Overall, CoGR-MoE remains robust on perception-heavy tasks but benefits more from structured prompting when deeper reasoning is required.

\begin{table*}[t]
\centering
\caption{Comparison of routing sharpness and routing variance on the \textbf{Perspective} categories.}
\label{tab:routing_perspective}
\begin{tabular}{l|c|cccc}
\toprule
\textbf{Methods} & \textbf{Overall} & Angle & Partial & Scope & Occlusion \\
\midrule
\multicolumn{6}{l}{\textit{Routing Sharpness}} \\
\midrule
MOE-LLaVA & 0.12 & 0.08 & 0.06 & 0.12 & 0.08 \\
CoGR-MoE (MOE-LLaVA) & 0.23 & 0.26 & 0.28 & 0.26 & 0.20 \\
\midrule
\multicolumn{6}{l}{\textit{Routing Variance}} \\
\midrule
MOE-LLaVA & 0.46 & 0.35 & 0.53 & 0.46 & 0.31 \\
CoGR-MoE (MOE-LLaVA) & 0.32 & 0.19 & 0.22 & 0.30 & 0.24 \\
\bottomrule
\end{tabular}
\end{table*}

\begin{table*}[t]
\centering
\caption{Comparison of routing sharpness and routing variance on the \textbf{Transformative} and \textbf{Other} categories.}
\label{tab:routing_transformative}
\begin{tabular}{l|cccc|c}
\toprule
\textbf{Methods} & Temporal & Deformation & Incomplete & Biological & Others \\
\midrule
\multicolumn{6}{l}{\textit{Routing Sharpness}} \\
\midrule
MOE-LLaVA & 0.21 & 0.13 & 0.07 & 0.13 & 0.18 \\
CoGR-MoE (MOE-LLaVA) & 0.20 & 0.20 & 0.24 & 0.18 & 0.20 \\
\midrule
\multicolumn{6}{l}{\textit{Routing Variance}} \\
\midrule
MOE-LLaVA & 0.65 & 0.43 & 0.24 & 0.46 & 0.72 \\
CoGR-MoE (MOE-LLaVA) & 0.47 & 0.33 & 0.21 & 0.35 & 0.58 \\
\bottomrule
\end{tabular}
\end{table*}

\subsection{Routing Sharpness and Variance Study}
To evaluate whether CoGR-MoE achieves more semantically aligned and decisive routing than MoE-LLaVA, routing is explicitly enhanced toward experts aligned with the correct semantic direction. Qwen-VL is not included here because its MoE layer uses 128 experts with Top-8 routing, under which the relative differences in gating distributions become extremely small and difficult to interpret. Expert-gating distributions are collected on MRAG-Bench, where each test sample is annotated with a semantic category (e.g., Angle, Occlusion, Deformation, Biological). Two complementary metrics are used:

Routing sharpness is defined as the difference between the average gating weight of the selected experts and that of the unselected experts. Formally, letting $T(x)$ denote the Top-$K$ experts selected for sample $x$:
\begin{equation}
\begin{aligned}
\text{Sharp}(x)
&=
\frac{1}{|T(x)|}\sum_{i \in T(x)} g_i(x)
\\
&\quad -
\frac{1}{E - |T(x)|}\sum_{i \notin T(x)} g_i(x),
\end{aligned}
\end{equation}
where $E$ is the total number of experts.
A higher value indicates stronger preference for the selected experts.

Routing variance is defined for each semantic category $c$ as:
\begin{equation}
\text{Var}_{c}
=
\frac{1}{E}
\sum_{i=1}^{E}
\operatorname{Var}_{x \in c}\!\big( g_i(x) \big),
\end{equation}

which quantifies the variability of expert-gating distributions among samples within the same semantic group, with lower variance indicating more stable and semantically consistent routing. To improve readability, we report routing variance scaled by a factor of 10 without affecting relative comparisons.

As shown in Table \ref{tab:routing_perspective} and Table \ref{tab:routing_transformative}, CoGR-MoE produces consistently sharper and more stable routing than MoE-LLaVA. For routing sharpness, CoGR-MoE on MoE-LLaVA increases the overall score from 0.12 to 0.23. Stronger improvements appear in categories such as Angle and Partial. For routing variance, CoGR-MoE substantially reduces intra-category variability. On MoE-LLaVA, the overall variance drops from 0.46 to 0.31, with larger reductions in Partial and Temporal.

The improvements in routing sharpness and variance stem from CoGR-MoE’s ability to explicitly steer the router toward experts aligned with the correct semantic direction.   Rather than relying on unconstrained gating logits, CoGR-MoE builds a semantic anchor from positive and negative cues and injects it as a learnable signal, nudging samples of the same semantic category toward a consistent subset of experts.   The probe-aligned contrastive objective further strengthens this direction, while uncertainty-aware weighting suppresses noisy gradients that could destabilize routing.

\section{Theoretical Analysis}
\subsection{Bounded Routing Perturbation}
\paragraph{Proposition.}
Let $z_{\text{base}} \in \mathbb{R}^E$ denote the original gating logits over $E$ experts, and let the semantic vector $b_{\text{CoGR}}$ satisfy $\|b_{\text{CoGR}}\|_\infty \le C$ for some constant $C>0$. For any bounded scalar $\lambda$, the semantic-guided routing
\begin{equation}
g^{T} = \mathrm{softmax}\bigl(z_{\text{base}} + \lambda b_{\text{CoGR}}\bigr)
\end{equation}
constitutes a bounded and Lipschitz-continuous perturbation of the original routing distribution $\mathrm{softmax}(z_{\text{base}})$. In particular, it does not introduce additional extrema nor destabilize the relative ordering of experts.

\paragraph{Proof.}
The softmax function is Lipschitz-continuous with respect to its input logits under the $\ell_\infty$ norm. That is, there exists a constant $L>0$ such that for any $z, z' \in \mathbb{R}^E$,
\begin{equation}
\|\mathrm{softmax}(z) - \mathrm{softmax}(z')\|_1
\le L \|z - z'\|_\infty.
\end{equation}
In the semantic-guided routing, the perturbation applied to the original logits is
\begin{equation}
(z_{\text{base}} + \lambda b_{\text{CoGR}}) - z_{\text{base}} = \lambda b_{\text{CoGR}}.
\end{equation}
Since $\|b_{\text{CoGR}}\|_\infty \le C$, we have
\begin{equation}
\|(z_{\text{base}} + \lambda b_{\text{CoGR}}) - z_{\text{base}}\|_\infty
\le \lambda C.
\end{equation}
Combining the above inequalities, the induced change in the routing distribution is bounded by $L \lambda C$. Therefore, the semantic enhancement introduces a controlled directional shift in logit space rather than an unbounded deformation, preserving routing stability and preventing expert collapse.

\subsection{Shared Top-$K$ Consistency}

Let $z_{\text{base}}$ denote the base gating logits produced by the router for an image--question pair $(I, Q)$, and let $\mathrm{TopK}$ denote the set of experts selected according to $z_{\text{base}}$. For each answer option $j$, option-specific logits over the shared Top-$K$ experts are defined as
\begin{equation}
\mathrm{logits}_{\text{top}}(j)
=
z_{\text{base}}[\mathrm{TopK}] + \lambda s_j[\mathrm{TopK}],
\label{eq:option_logits}
\end{equation}
where $s_j$ denotes the semantic signal derived from the option text. The corresponding gating distribution is given by
\begin{equation}
g_j = \mathrm{softmax}\bigl(\mathrm{logits}_{\text{top}}(j)\bigr),
\end{equation}
and the aggregated representation for option $j$ is computed as
\begin{equation}
\tilde{h}_j = \sum_{i \in \mathrm{TopK}} g_j(i)\, h_i.
\label{eq:option_aggregation}
\end{equation}

Since the Top-$K$ expert set is determined solely by $z_{\text{base}}$, it is identical for all answer options and independent of the option index $j$. Consequently, routing decisions are shared across options and depend only on the input $(I, Q)$. $s_j$ influences the model exclusively through reweighting within this shared expert set, without altering the routing outcome.

As all option representations $\tilde{h}_j$ are linear combinations of the same expert outputs, they lie in a common expert subspace. This design ensures comparability across options and prevents option-conditional routing drift. In contrast, allowing each option to independently select its own Top-$K$ experts would entangle routing with option identity, leading to duplicated expert computation and destabilized expert specialization.

\subsection{Bounded Gradient Reweighting}

\paragraph{Proposition.}
Let the main training objective be defined as
\begin{equation}
L_{{main}} = \sum_j \frac{1}{1+{unc}_j}\,
\mathrm{CE}(\mathrm{score}(j), y_j).
\end{equation}
where $\mathrm{CE}_j(\theta)$ denotes the cross-entropy loss associated with option $j$, and ${unc}_j \ge 0$ is the corresponding uncertainty-based weight. This formulation induces a bounded rescaling of per-option gradients and does not introduce uncontrolled gradient amplification.

\paragraph{Proof.}
Taking the gradient of $L_{\text{main}}$ with respect to model parameters $\theta$ yields
\begin{equation}
\nabla_\theta L_{{main}}
=
\sum_j {unc}_j \, \nabla_\theta \mathrm{CE}_j(\theta).
\end{equation}
Applying the triangle inequality, we obtain
\begin{equation}
\begin{aligned}
\|\nabla_\theta L_{{main}}\|
&\le
\sum_j {unc}_j \, \|\nabla_\theta \mathrm{CE}_j(\theta)\| \\
&\le
\bigl(\max_j {unc}_j\bigr)
\sum_j \|\nabla_\theta \mathrm{CE}_j(\theta)\|.
\end{aligned}
\end{equation}

Therefore, uncertainty-aware weighting amounts to a bounded linear rescaling of per-option gradients. As long as ${unc}_j$ is bounded, the overall gradient magnitude remains controlled, ensuring that uncertainty-based reweighting does not destabilize optimization but only modulates the relative influence of different options during training.

\subsection{Contrastive-Alignment Consistency}

\paragraph{Proposition.}
Let the semantic alignment metric be defined as
\begin{equation}
\mathrm{Sim} = \cos(h_{\mathrm{TopK}}, s_a),
\end{equation}
where $h_{\mathrm{TopK}}$ denotes the aggregated representation of the routed Top-$K$ experts and $s_a$ is the semantic direction corresponding to the correct answer.
Consider the cue-guided contrastive objective used in Eq.~(9), which encourages the aggregated representation of the correct option to align with positive semantic cues while pushing incorrect option representations away from negative cues.
Then, minimizing the contrastive loss induces gradient updates on the shared expert representations that are directionally consistent with increasing the semantic alignment metric $\mathrm{Sim}$.

\paragraph{Proof.}
The cosine similarity measures the angular alignment between two vectors in representation space.
Without loss of generality, we assume $\|s_a\| = 1$, since normalization does not affect directional analysis.
For a representation $h$, the cosine similarity with respect to $s_a$ can be written as
\begin{equation}
\cos(h, s_a) = \frac{h^\top s_a}{\|h\|}.
\end{equation}
Taking the gradient with respect to $h$ yields
\begin{equation}
\nabla_h \cos(h, s_a)
= \frac{s_a}{\|h\|}
- \cos(h, s_a)\frac{h}{\|h\|^2}.
\end{equation}
The first term promotes rotation of $h$ toward the semantic direction $s_a$, while the second term removes the component aligned with $h$ itself, preventing trivial norm inflation.
As a result, minimizing a negative cosine similarity term $-\cos(h, s_a)$ induces gradient updates that rotate the representation toward $s_a$ while keeping its norm controlled.

In our setting, the contrastive loss in Eq.~(9) is defined over option-specific aggregated representations, including the correct-option representation $\tilde{h}_{\mathrm{correct}}$ and a pooled incorrect-option representation $\tilde{h}_{\mathrm{wrong}}$.
Both representations are formed by reweighting the same shared Top-$K$ expert outputs.
Consequently, gradients from the contrastive objective propagate to the shared expert representations that constitute $h_{\mathrm{TopK}}$.

The positive term in the contrastive loss aligns $\tilde{h}_{\mathrm{correct}}$ with the positive semantic cues, which are constructed to be consistent with the semantic direction $s_a$.
This induces gradient updates on the selected experts that rotate their representations toward $s_a$.
Meanwhile, the negative term discourages alignment of incorrect-option representations with incompatible semantic cues, further suppressing expert activations that conflict with the correct semantic direction.

Therefore, although the contrastive objective does not explicitly maximize $\mathrm{Sim}$, its gradient induces a consistent rotational pressure on the shared expert representations toward the correct semantic direction.
As training proceeds, this alignment-consistent gradient flow increases the expected cosine similarity between $h_{\mathrm{TopK}}$ and $s_a$, corresponding to an increase in the semantic alignment metric $\mathrm{Sim}$.

\section{Prompt Listing}

\subsection{Full Prompt}
\begin{Verbatim}[breaklines=true,fontsize=\small]
You are a visual evidence aligner.

Given an image I, a question Q, and a set of candidate answer options,
your goal is to generate decision-level semantic cues for each option.

For each option a, produce two sets of cues:
- positive-cues: observable visual evidence that should be present if a is correct.
- negative-cues: observable evidence that would contradict a if present.

Before generating cues, identify the main task type(s) involved in the question
(multiple types may apply), and tailor the cues accordingly.

Task categories include:
- Perception
- Counting
- Spatial
- OCR / Text
- Commonsense & Knowledge
- Reasoning

[Category-Specific Guidance]

Perception:
Focus on directly observable visual attributes, such as object parts, colors, materials,
textures, or the presence or absence of specific entities. Cues should be localized
and visually verifiable.

Counting:
Focus on explicit countable instances or visual anchors that support a specific quantity.
Negative cues should highlight visible evidence that contradicts an incorrect count.

Spatial:
Focus on relative spatial relationships between entities, such as left/right, above/below,
inside/outside, distance, orientation, or occlusion relationships.

OCR / Text:
Focus on visible text or symbols in the image. Positive cues should describe readable text
or character patterns together with coarse location information. Avoid hallucinating text;
if the text is uncertain, indicate low confidence rather than fabricating content.

Commonsense & Knowledge:
Use only cues that are directly supported by observable visual evidence.
Do not rely on external knowledge as decisive evidence when generating cues.

Reasoning:
Describe minimal sets of observable premises together with a short rule or relation
that supports the option. Avoid long chains of reasoning and do not introduce
external knowledge beyond what is visible or stated in the question.

[General Principles]
- Cues should be concrete, specific, and verifiable from the image.
- Prefer minimal but discriminative evidence.
- Avoid subjective or non-observable attributes.
- If evidence is weak or uncertain, reflect this by lowering confidence rather than inventing details.

Return the positive-cues and negative-cues for each option in a structured format.
\end{Verbatim}

\subsection{Minimal Prompt}
\begin{Verbatim}[breaklines=true,fontsize=\small]
You are given an image, a question, and multiple candidate answer options.

For each option a, generate:
- positive-cues: a small set of visible cues that must be present in the image if the option were correct.
- negative-cues: visible cues that, if present, would contradict the option.

Cues should be concrete, observable, and directly verifiable from the image.
Do not rely on external knowledge or subjective descriptions.
If evidence is weak or uncertain, lower the confidence rather than inventing details.

Output a concise JSON object containing, for each option, its positive-cues and negative-cues.
\end{Verbatim}

\end{document}